\documentclass[conference]{IEEEtran}
\IEEEoverridecommandlockouts
\usepackage[utf8]{inputenc}
\usepackage[numbers]{natbib}
\usepackage{amsmath,amssymb,amsfonts}
\usepackage{algorithmic}
\usepackage{graphicx}
\usepackage{textcomp}
\usepackage{booktabs}
\def\BibTeX{{\rm B\kern-.05em{\sc i\kern-.025em b}\kern-.08em
    T\kern-.1667em\lower.7ex\hbox{E}\kern-.125emX}}

\begin{document}

\title{A Convolutional Network for Sleep Stages Classification\\
\thanks{* This research was funded by the Xunta de Galicia (ED431G/01) and the European Union (ERDF).}
}
\author{
    \IEEEauthorblockN{Isaac Fernández-Varela\IEEEauthorrefmark{1}, Elena Hernández-Pereira\IEEEauthorrefmark{1}, Diego Alvarez-Estevez\IEEEauthorrefmark{2} and Vicente Moret-Bonillo\IEEEauthorrefmark{1}}
    \IEEEauthorblockA{\IEEEauthorrefmark{1}\textit{CITIC} \\
    \textit{Universidade da Coruña}\\
    A Coruña, España \\
    (isaac.fvarela, elena.hernandez, vicente.moret)@udc.es}
    \IEEEauthorblockA{\IEEEauthorrefmark{2}\textit{Sleep Center \& Clinical Neurophysiology} \\
    \textit{Haaglanden Medisch Centrum}\\
    The Hague, The Netherlands \\
    diego.alvarez@udc.es}
}

\maketitle

\begin{abstract}
Sleep stages classification is a crucial task in the context of sleep studies. It involves the simultaneous analysis of multiple signals recorded during sleep. However, it is complex and tedious, and even the trained expert can spend several hours scoring a single night recording. Multiple automatic methods have tried to solve these problems in the past, most of them by classifying a feature vector that is engineered for a specific dataset. In this work, we avoid this bias using a deep learning model that learns relevant features without human intervention. Particularly, we propose an ensemble of 5 convolutional networks that achieves a kappa index of $0.83$ when classifying a dataset of 500 sleep recordings. 

\end{abstract}


\section{Introduction}
Sleep disorders affect a major part of the population. As an example, 20\% of the Spanish adults suffer insomnia, and between 12\% and 15\% daytime sleepiness~\citep{Ohayon2010, Marin1997}. Good sleep is essential for a healthy life, and the adverse consequences of restless nights have been extensively reported~\citep{Colten2006}. To evaluate the sleep function, and to help the diagnosis of sleep disorders, it is important to know the sequence of sleep stages that the patient goes through the night. 

The most common technique to monitor the sleep function is the polysomnogram (PSG), which involves recording of the patient's biosignals during sleep, including various pneumological, electrophisiological, and contextual information. This is an expensive test, uncomfortable for the patient, and for which interpretation of the results is difficult due to the complexity of the data involved. An usual way to summarize the sleep information contained in the PSG is the derivation of the hypnogram, an ordered representation of the sleep stages evolution.

The current gold standard for the building the hypnogram is the \textit{American Academy of Sleep Medicine} (AASM)~\citep{Berry2017} guide for the identification of sleep stages and of their associated events (e.g. EEG arousals, limb movements, and cardiac or respiratory events). This guide identifies five sleep stages: Awake (W), Rapid Eye Movements (REM), and 3 non-REM phases (N1, N2, and N3). Correct identification of the sleep stages and construction of the hypnogram is of fundamental importance to achieve a good diagnosis, allowing the clinician to focus efforts in the therapy. Such a task implies the analysis of huge amounts of data and expert knowledge~\citep{Fernandez-Leal2017}. Moreover, even following the guidelines, inter-expert agreement usually remains below the 90\%. For example, \citet{Stepnowsky2013} studied the agreement between two experts finding kappa index values between $0.48$ and $0.89$. Similarly, \citet{Wang2015} found values between $0.72$ and $0.85$. Furthermore, agreement is worse for some particular stages, usually being stage N1 the one with the highest disagreement. 

%El estándar de oro para la construcción del hipnograma es la guía de la \textit{American Academy of Sleep Medicine} (AASM)~\citep{Berry2017} para la identificación de fases del sueño y sus eventos asociados, los despertares, los movimientos y los eventos cardíacos y respiratorios. Dicha guía identifica cinco fases de sueño: Despierto (\textit{Awake}, W), movimientos oculares rápidos (\textit{Rapid Eye Movements}, REM), y tres fases de sueño lento o no REM (N1, N2 y N3). Un hipnograma construido correctamente facilita encontrar problemas y diagnosticar trastornos del sueño, permitiendo enfocar el tiempo en la terapia. Su construcción implica analizar una gran cantidad de información y conocimiento~\citep{Fernandez-Leal2017}. Además, pese a las guías el acuerdo entre expertos es usualmente inferior al 90\%. Por ejemplo, \citet{Stepnowsky2013} estudiaron el acuerdo entre dos expertos obteniendo índices kappa entre 0,48 y 0,89. De manera similar, \citet{Wang2015} obtuvieron índices entre 0,72 y 0,85. A mayores, el acuerdo también es menor para fases concretas, siendo la fase N1 la que obtiene los peores resultados. 

All given, automatic methods for sleep stages classification are needed. Most of these methods follow a two step approach. First, feature extraction takes place, usually with features hand tailored for a specific dataset. Then, feature vectors are built to train a classifier and predict the sleep stages. While some authors have used a single signal channel as reference (usually the EEG), other approaches have extracted features using several channels, building input vectors of various elements. At this respect usually features from the electrooculogram (EOG) or electromiogram (EMG) are added to those of the EEG, as recommended by the AASM guidelines. Often features are extracted either from to the time or from the frequency domain. 

Among the methods following this 2-step approach we find:
\citet{Fraiwan2012} use a random forest to classify features both from the time-frequency domain and Renyi's entropy;
\citet{Liang2016} measure entropy with different scales obtaining autoregresive features which classify using a linear discriminant;
\citet{Hassan2016a}, apply wavelet transformations for feature extraction and use a random forest technique for the classification step. 
\citet{Sharma2017}, compare several classifiers for iterative filters analysing a single EEG channel;
\citet{Koley2012}, train a support vector machine (SVM) with frequency, time and non-linear features extracted from a single EEG channel;
\citet{Lajnef2015}, base their approach on multiple signals building a decision tree upon several SVMs;
\citet{Huang2014}, study power spectral density of 2 EEG channels classifying frequency features with a modified SVM;
Finally, \citet{Gunes2010}, also analyse power spectral density while classifying with a nearest neighbours algorithm.

The approach consisting in solving the sleep staging classification problem using handcrafted feature extraction induces biases due to the design of features based on one specific database. Thus, the aforementioned solutions usually do not generalize well, specially given the nature of PSG recordings, where variability effects are introduced due to several factors, including patient, hardware or scoring differences. 

One alternative option to solve this problem is the use of methods than learn directly from the raw data, therefore avoiding the human bias. In this sense, deep learning represents a  natural approach, as it demonstrated improvements against traditional methods in multiple general fields, including in particular, the medical diagnosis~\citep{Esteva2017a, Gulshan2016}.

Some works have already explored solutions with different deep learning models:
\citet{Langkvist2012}, used deep belief networks learning a probabilistic representation of preprocecessed signals from PSG inputs;
\citet{Tsinalis2016}, still followed the 2 step approach, but with convolutional networks for classification. In other work, the same authors~\citep{Tsinalis2016a} relied on a stack of \textit{sparse autoencoders};
\citet{Supratak2017}, performed classification from the raw signals with a bidirectional recurrent neural network;
\citet{Biswal2017}, compared a recurrent network against different models, although all were trained with features instead of the raw signal;
Finally, \citet{Sors2018} also used a convolutional neural network using one single EEG channel as reference.

In this work we use deep learning to classify sleep stages with a convolutional neural network that learns the relevant features for each stage. Following the AASM guidelines we use multiple signals; namely, two EEG, one EMG, and two (left and right) EOG channels. Moreover, signals are filtered in the first place, to reduce noise and remove artifacts. 

\section{Materials}
\label{section:materials}

Design and analysis of the presented model was carried out using PSG recordings from real patients. These recordings belong to the Sleep Heart Health Study (SHHS)~\citep{Quan1997}, a database offered by the Case Western University, originated from a cohort study involving multiple centers directed by the National Heart Lung and Blood Institute, with the goal of determining the cardiovascular consequences of respiratory related sleep disorders. 

Each recording contains annotations for different events performed by clinical experts following the procedures described in ~\cite{SHHS2002}. All recordings were anonymized and blind scored. The montage for the signals acquisition included two EEG derivations (C4A2 and C4A1), left and right EOGs, chin EMG, and modified lead-II electrocardiogram (ECG). EEG, EOG, and EMG were sampled at 125 Hz whereas EOG were sampled at 50 Hz. All signals were filtered during acquisition with a high pass filter at 0.15 Hz.

From this database three different datasets were selected to train, validate and test our model. Training dataset included 400 recordings, validation 100, and test 500. The length of the training recordings is matched (limiting each to a total of 7 randomly selected hours) to facilitate the coding and the training of the algorithm. Finally, our training dataset contained $288.000$ $30-s$ epoch samples, the validation dataset $119.121$ and the test dataset $606.981$. Recordings were selected randomly, including those with high levels of noise or artifacts. 

The distribution for the different classes, both for the complete dataset as for each individual recording is shown in Table~\ref{table:datasets}. This table shows how unbalanced the datasets are, being W the most represented class (about 38\% of the samples), although with a similar proportion to N2 (around 36\%). On the contrary, class N1 is only represented in 3\% of the classes
It is also interesting to notice how some recordings do not contain samples for some of the classes, and how much the distribution differs between the recordings. For example, in the test dataset, whereas a particular recording contains a 7.10\% of samples for class N2, another goes up to a 83.43\%. Moreover, these are the two important problems when trying to develop an automatic sleep staging classifier: 1) the class unbalance and 2) the differences between individual recordings.

\begin{table*}[htp]
\caption{Distribution of the different classes in the training, validation, and test datasets.}
\label{table:datasets}
\centering
\begin{tabular}{@{}llrrrrrr@{}}
\toprule
           &                   		& W 		& N1		& N2		& N3		& REM		& Total		\\ 
\midrule
\textit{Training dataset} & \hspace{0.4cm} Total				& 187.513 	& 17.283 	& 172.451 	& 44.454 	& 62.168 	& 483.869 	\\
& \hspace{0.4cm} Proportion           & 38,75 \%  & 3,57 \%   & 35,64 \%  & 9,19 \%   & 12,85 \%  & 100 \% 	\\
& \hspace{0.4cm} Min in single record 	& 8,20 \%   & 0,00 \%	& 12,59 \%	& 0,00 \%	& 0,00 \% 	&           \\
& \hspace{0.4cm} Max in single record	& 71,61 \%  & 13,75 \% 	& 68,65 \% 	& 33,43 \%	& 26,58 \% 	&           \\
\midrule
\textit{Validation dataset} & \hspace{0.4cm} Total                & 43.742 	& 3.963 	& 43.510 	& 12.900 	& 15.006 	& 119.121	\\
& \hspace{0.4cm} Proportion           & 36,72 \%	& 3,33 \%   & 36,53 \%  & 10,83 \%  & 12,60 \%  & 100 \% 	\\
& \hspace{0.4cm} Min in single record 	& 11,21 \%	& 0,29 \% 	& 12,38 \%	& 0,00 \%	& 0,00 \% 	&			\\
& \hspace{0.4cm} Max in single record 	& 76,79 \% 	& 17,08 \% 	& 60,09 \% 	& 30,16 \% 	& 23,68 \% 	&			\\
\midrule
\textit{Test dataset} & \hspace{0.4cm} Total                & 231.707 	& 19.769 	& 217.246 	& 61.281 	& 76.978 	& 606.981 	\\
& \hspace{0.4cm} Proportion			& 37,77 \%  & 3,26 \%	& 35,96 \%  & 10,25 \%  & 12,75 \%  & 100 \%	\\
& \hspace{0.4cm} Min in single dataset 	& 7,75 \% 	& 0,00 \% 	& 7,10 \% 	& 0,00 \% 	& 0,00 \% 	&			\\ 
& \hspace{0.4cm} Max in single dataset 	& 76,53 \%  & 16,93 \% 	& 83,43 \%  & 43,82 \% 	& 31,11 \% 	&			\\ 
\bottomrule
\end{tabular}
\end{table*}

\section{Methods}

\subsection{Signal filtering}

Signals are preprocessed to reduce noise and remove common artifacts. Both operations are typically applied in previous works before feature extraction. 

The first of the two filters used to reduce noise is a Notch filter centered at 60 Hz to remove mains interference. This filter is applied to those signals with a sampling rate higher than 60 Hz: EEG and EMG. The second one removes DC component and frequencies not related with muscular movements from the EMG, applying a high pass at 15 Hz. 

Regarding artifacts, most of then happen during particular short time periods, making it difficult even their detection. 
However, ECG artifacts, caused by the heart beat interference, are common and constant through the whole signals. 
We can remove this kind of artifact with an adaptive filter. 
To do so, we first obtained the beat series following a standard QRS detection algorithm~\citep{Afonso1999}. 
Then, we studied the signal quality to asses which intervals could be safely included in the construction of the adaptive filter. 
Finally, during the intervals with enough signal quality, we applied and updated the filter template to remove the artifacts. 
More information about this process can be found in \citet{Fernandez-Varela2017}.

\subsection{Convolutional network}

Sleep stages classification is usually carried out with 30 s windows called epochs. Analyzing several features from each epoch, clinicians score the corresponding sleep stage.

\begin{figure}
\centering
\includegraphics[width=\linewidth]{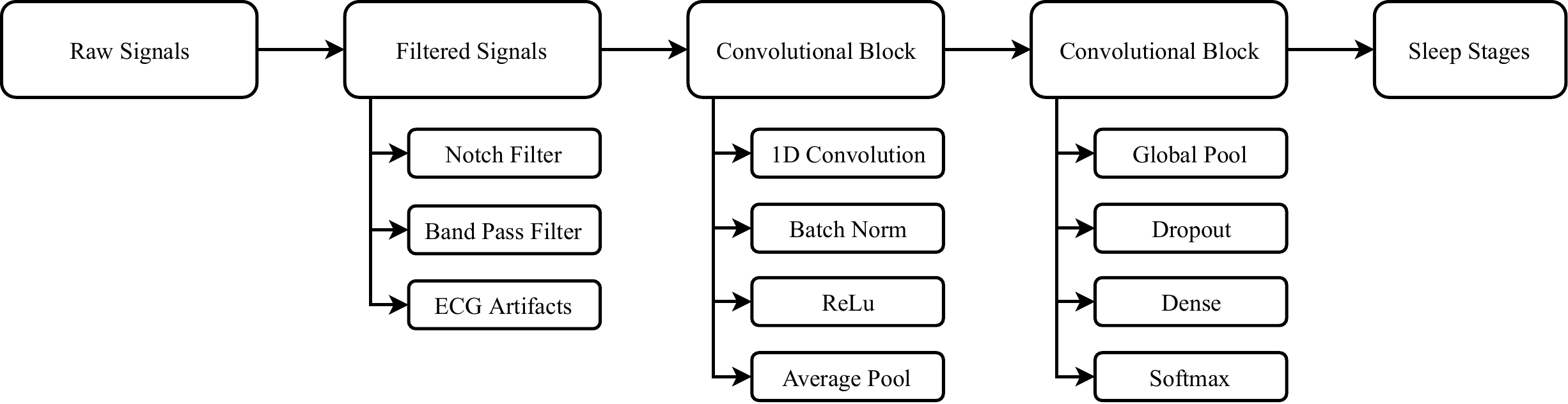}
\caption{Proposed convolutional neural network}
\label{figure:network}
\end{figure}

A convolutional neural network~\citep{LeCun1989} is a feedforward network solving the limitations of the multilayer perceptron with a weight sharing architecture. Basically, it applies a convolution operation over the input, limiting the number of parameters. Thus, it allows the construction of deeper networks that are better at recognizing complex features.
The proposed network is represented in Figure~\ref{figure:network}.

The input to the convolutional network is the set of signals (2 EEG channels, EMG, and both EOGs). Each input pattern corresponds to a 30 s epoch window. As the signals are sampled at different rates (aforementioned in Section~\ref{section:materials}) we upsampled those with sampling rates lower than 125 Hz. We avoided downsampling to 50 Hz because it would mean loosing high frequencies in the EEG that should contain important information from a clinical perspective. Moreover, we also discarded padding because the approach cannot be easily generalized to other datasets with different sampling rates. This way, each input to the network is a matrix with a dimension of $3750 \times 5$. Each signal was normalized with mean 0 and deviation 1, using the mean and deviation obtained from all the respective signals in the training dataset. When we tried other normalizations with lower granularity, our training did not converge. 
The convolutional block shown in Figure~\ref{figure:network} is a stack of four layers including a 1D convolution that preservers the input dimension (with padding), a batch normalization layer~\citep{Ioffe2015} to improve regularization, ReLu~\citep{Nair2010} activation, and an average pool that reduces dimension by a factor of 2. By using 1D convolution we avoided imposing a spatial structure between our signals that is unknown a priori. This stack was repeated $n$ times, being $n$ an hyperparamenter with a value selected during experimentation. All layers were configured with the same kernel size but the number of filters for layer $i$ is twice the number of filters for layer $i - 1$. The selection value of $n$, the kernel size and the number of filters for the first layers is explained in the following Section, together with the remaining hyperparameters. 

The output of the last convolutional block, after adjusting dimensions with a global pooling and applying dropout, is used as input for a dense layer with a softmax activation. This layer returns the probability for each sleep stage given the initial input. As usual, the final predicted class is set to the output showing the highest probability.

To train the network we used Adam optimizer~\citep{Kingma2014} and a batch size of 64. This batch size was limited by our hardware. The learning rate was configured whereas both betas are left with the default values. Training ends using early stopping by monitoring the validation loss with a patience of 10 epochs. To limit the impact of class unbalance, we used weighted cross entropy as the cost function, where weights were obtained using the training dataset. 

\subsection{Hyperparameter optimization}

A good selection of hyperparameters can mean the success of a deep learning model. The difficulty when selecting the best hyperparameters is not only to achieve the best performance, but doing it while at the same time minimizing the cost, either the economical or the computational cost. 

In this work we relied on a Tree-structured Parzen Estimator (TPE) that has shown better performance than other methods~\citep{Bergstra2011, Bergstra2013}. TPE is a sequential models based optimization. This kind of methods builds models sequentially to approximate the performance of hyperparameters selection based on historical results, and then chooses new hyperparameters that are checked with the model. Particularly, TPE uses two distributions $P(x|y)$ and $P(y)$ where $x$ represents the hyperparameters and $y$ the expected performance. The expected improvement (EI) is optimized according to the following equation: 

$$ EI_{y^*}(x) = \int_{-\infty}^{y^*} (y^* - y)\frac{P(x|y)P(y)}{P(x)}$$

where $y^*$ is a quantil $\gamma$ of the observed values $y$ such as $p(y < y^*) = \gamma $.

We used TPE to select the best values for the following hyperparameters related with the convolutional network: the number of convolutional blocks, kernel size for the 1D convolutions, and the number of filters for the first convolutional block. Moreover, there is also a relationship between the number of blocks and the number of initial filters. Given our hardware restrictions, we did not add blocks that would have more than 1024 filters. We also used TPE to select the learning rate. The distributions for the random values of each of these hyperparameters are summarized in Table~\ref{table:hyperparameter_distribution}.

\begin{table}[htp]
\caption{Distributions for the hyperparameters}
\label{table:hyperparameter_distribution}
\centering
\begin{tabular}{@{}ll@{}}
\toprule
Hyperparameter       & Distribution \\ 
\midrule
Convolutional Blocks    & Uniform between 1 and 10			\\
Kernel Size         	& Uniform between 3 and 50      	\\
First Block Filters     & Choice between 8, 16, 32 o 64	\\
Learning Rate           & Log-uniform between -10 and -1	\\
\bottomrule
\end{tabular}
\end{table}

To reduce the computational time for the hyperparameter selection we used a subset from the training set in order to train, validate, and test the different models. This subset contained 250 recordings where 20 were used for validation during training, and 50 to test each model. In total, we tried 50 different hyperparameter configurations, using the kappa index obtained with the test set as the criterion to select the best one. 
\subsection{Performance}

The performance of the models was evaluated using the following metrics:

\begin{itemize}
\item \textbf{Precision,} the fraction between true positives and the predicted positives.

\item \textbf{Sensitivity}, the fraction between true positives and the samples belonging to that class. 

\item \textbf{\textit{F1 score}}, harmonic mean between precision and sensitivity.

\item \textbf{Kappa}, agreement measure between two classifiers that takes into account the chances of random agreement. Perfect agreement gets a value of 1, and by chance a value of 0.

\end{itemize}

\section{Results}
\begin{figure}
\centering
\includegraphics[width=\linewidth]{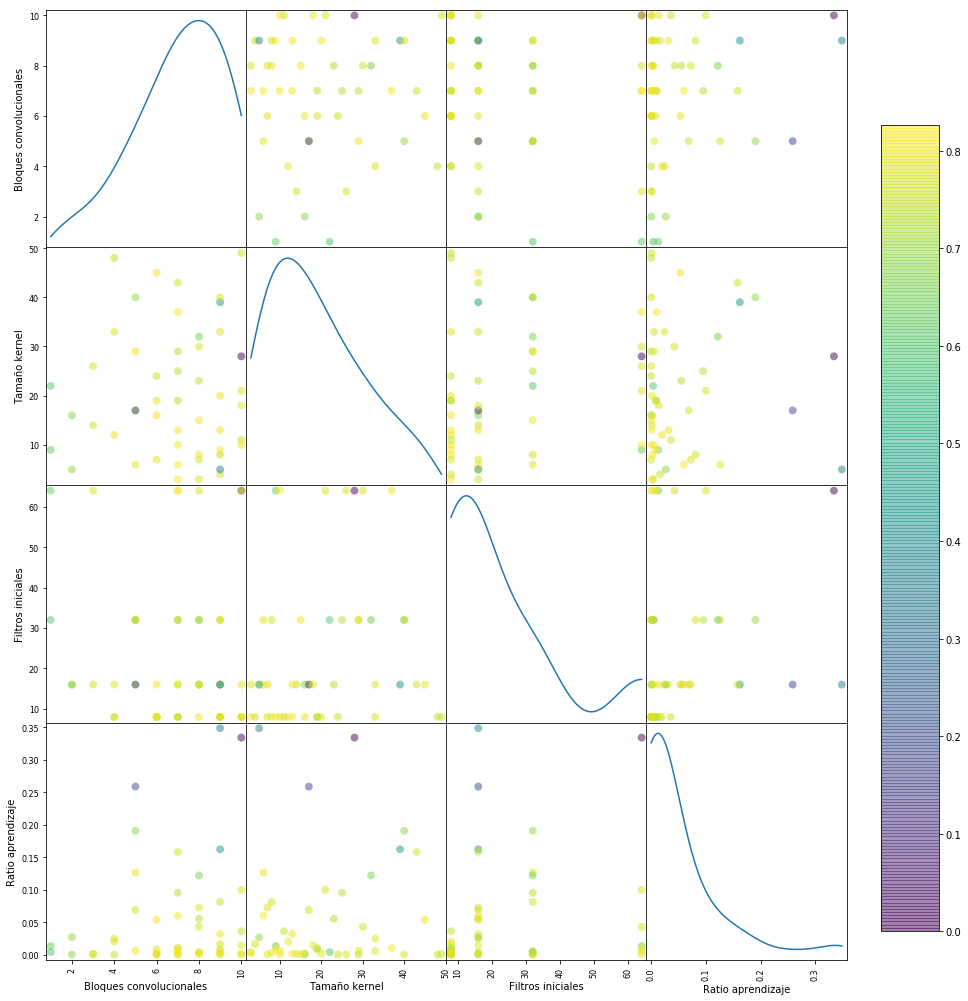}
\caption{Dispersion graph with the different configurations of hyperparameters. Each point color represents the kappa index for the model with the values for the hyperparameters represented in the axes. Diagonal represents the distribution for the values tried for a particular hyperparameter.}
\label{figure:trials}
\end{figure}

Before focusing on the results achieved with the final model, performance of the different models evaluated during the hyperparameters search is shown in Figure~\ref{figure:trials}. Data in the figure suggest a clear trend toward low learning rates to ensure convergence. 

To improve the results obtained by a single model we used an ensemble. Thereby, several models classify the same input, and the final decision is taken using the majority vote. In this case, we selected the 5 best models obtained during the hyperparameter selection. Values for the hyperparameters for each of those models are shown in Table~\ref{table:hyperparameter_distribution}.

\begin{table*}[htp]
\caption{Hyperparameters for the 5 models with the best kappa index}
\label{table:hyperparameter}
\centering
\begin{tabular}{@{}lccccc@{}}
\toprule
Parameter               & Model 1 & Model 2 & Model 3 & Model 4 & Model 5 \\ 
\midrule
Convolutional blocks 	& 7						& 9 & 7 & 7 & 7 \\
Kernel size      		& 6      				& 9 & 13 & 3 & 10\\
Initial filters         & 16					& 8 & 8 & 8 & 64 \\
Learning rate        	& $5,99 \times 10^{-2}$ & $9,00 \times 10^ {-3}$ & $ 1,45 \times 10^ {-3}$ & $ 1,91  \times 10^ {-3} $ & $ 5,49  \times 10^ {-3} $ \\
\bottomrule
\end{tabular}
\end{table*}

Results obtained with the ensemble using the test set are shown in Table~\ref{table:inputs_comparison}. The best classification was achieved for class W, with values near to $0.95$ for the precision, sensitivity and F1 score; then, classes N2, N3, and REM showed similar results, specially if we compare the F1 score, although sensitivity for N3 was lower (thus, precision was higher). Lastly, results regarding the the classification of class N1 were rather low, not even achieving a F1 score of $0.3$. However, N1 is typically the most difficult class to predict, showing the highest disagreement also among trained experts. 

\begin{table}[ht!]
\caption{Performance measures for the classification of the test dataset using the ensemble with the 5 selected models.}
\label{table:inputs_comparison}
\centering
%\resizebox{\textwidth}{!}{%
\begin{tabular}{@{}lrrr@{}}
\toprule

\textbf{Stage} & Precision    & Sensitivity   & \textit{F1 score}   \\
\midrule
W     & 0,94         & 0,96     & 0,95     \\
N1    & 0,39         & 0,21     & 0,27     \\
N2    & 0,87         & 0,89     & 0,88     \\
N3    & 0,92         & 0,77     & 0,84     \\
REM   & 0,82         & 0,90     & 0,86     \\ 
\midrule
\textbf{Average} & 0,78 & 0,75 & 0,76 \\
\bottomrule
\end{tabular}
\end{table}

\begin{figure}[ht!]
\centering
\includegraphics[width=.8\linewidth]{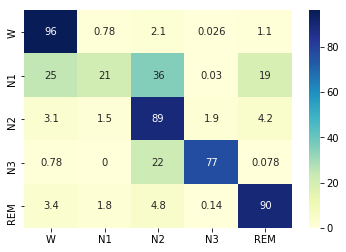}
\caption{Confusion matrix for the classification of the test dataset using the ensemble with the 5 selected models.}
\label{figure:confusion_matrix}
\end{figure}

The confusion matrix obtained with the ensemble is shown in Figure~\ref{figure:confusion_matrix}, where we can verify how most of the N1 samples are misclassified, specially towards class N2. Also, although in a smaller proportion, whenever there is a classification error it tends to be misclassifying as N2. 

\section{Discussion and Conclusions}

In this work we present an ensemble of convolutional networks for the classification of sleep stages. Sleep staging is a time consuming task, nevertheless critical for a good diagnosis of sleep disorders. Most of the automatic methods reported so far are based on human engineered features, designed for a particular dataset. Thus, it is difficult to find a method that generalizes correctly to other datasets. To solve this problem we propose the use of a convolutional network that self learns the relevant features for the classification, avoiding human biases. 

An important aspect for the success or failure of convolutional methods is the correct choice of the hyperparameters. In this paper, we experimented with 4 hyperparameters, optimizing their values with a tree-structured parzen estimator, trying 50 different configurations. 

Our ensemble, built from the best 5 hyperparameters configurations, achieved an average precision, sensitivity, and F1 score of $0,78$, $0,75$ y $0,76$ respectively, with a kappa index value of $0.83$. Although globally our results are acceptable, our solution has shown problems for the classification of class N1. Also, in the event of misclassification, a trend has been noticed towards class N2. 

Comparison of our results against similar works is difficult given the lack of standardization, both as with regard to the chosen datasets, as well as in the procedures for the evaluation process. In Table~\ref{table:previous_works} we show results from previous works, limiting to those that report values separately for each class. As it can be seen, our kappa index is the highest, although it is not the case for the F1 score. According to the F1 score, and apart from class W, some works are able to achieve better classification for the remaining classes. However, the values that we obtained are competitive, excluding class N1, although it is clear from all the results, that this is the most difficult class. Taking as reference the only work showing results with a similar dataset~\citep{Sors2018}, our kappa index and F1 score for W class are higher, with similar values for N2, N3, and REM but lower for class N1. 

\begin{table*}[ht!]
\caption{Comparison against previous works.}
\label{table:previous_works}
\centering
\begin{tabular}{llrrrrrr}
\toprule
\textbf{Work}			 			& \textbf{Database} & \multicolumn{1}{c}{\textbf{Kappa}} & \multicolumn{5}{c}{\textbf{F1 score}} 	    \\
				 			& &   	&      \textbf{W} &   \textbf{N1} &   \textbf{N2} &   \textbf{N3} &  \textbf{REM} \\
\midrule
\citet{Biswal2017}          & Massachusetts General Hospital, 1000 recordings &  0,77 &   0,81 & \textbf{0,70} & 0,77 & 0,83 & \textbf{0,92} \\
\citet{Langkvist2012}       & St Vicent's University Hospital, 25 recordings &  0,63 &   0,73 & 0,44 & 0,65 & \textbf{0,86} & 0,80 \\
\citet{Sors2018}            & SHHS, 1730 recordings &  0,81 &   0,91 & 0,43 & 0,88 & 0,85 & 0,85 \\
\citet{Supratak2017}     	& MASS dataset, 62 recordings &  0,80 &   0,87 & 0,60 & \textbf{0,90} & 0,82 & 0,89 \\
\citet{Supratak2017} 		& SleepEDF, 20 recordings &  0,76 &   0,85 & 0,47 & 0,86 & 0,85 & 0,82 \\
\citet{Tsinalis2016}    	& SleepEDF, 39 recordings &  0,71 &   0,72 & 0,47 & 0,85 & 0,84 & 0,81 \\
\citet{Tsinalis2016a}   	& SleepEDF, 39 recordings &  0,66 &   0,67 & 0,44 & 0,81 & 0,85 & 0,76 \\
\midrule
This work 					& SHHS, 500 recordings &  \textbf{0,83} &   \textbf{0,95} & 0,27 & 0,88 & 0,84 & 0,86 \\
\bottomrule
\end{tabular}
\end{table*}

Our results are promising and the chosen method should be easily adaptable to other datasets, specially if we can train the model for the different dataset. Moreover, training it with more than one dataset should improve generalization, avoiding biases for a single dataset. 

To improve our result it is necessary to understand why and how the network is classifying. Also, it would be interesting to add memory to the model using recurrent networks, as the classification of some inputs, following the clinical definition, depends as well on the status of the neighbouring epochs. 
\small
\bibliographystyle{IEEEtranN}
\bibliography{library.bib}

\end{document}